**Transformers and Cortical Waves: Encoders for Pulling In Context Across Time**

Lyle Muller[1,2], Patricia S. Churchland[3], Terrence J. Sejnowski[4,5]
[1]Department of Mathematics, Western University
[2]Fields Lab for Network Science, Fields Institute
[3]Department of Philosophy, University of California at San Diego
[4]Computational Neurobiology Laboratory, Salk Institute for Biological Studies
[5]Department of Neurobiology, University of California at San Diego

**Correspondence:** lmuller2@uwo.ca, pschurchland@ucsd.edu, terry@salk.edu

**Abstract:** The capabilities of transformer networks such as ChatGPT and other Large Language Models (LLMs) have captured the world's attention. The crucial computational mechanism underlying their performance relies on transforming a complete input sequence – for example, all the words in a sentence – into a long "encoding vector" that allows transformers to learn long-range temporal dependencies in naturalistic sequences. Specifically, "self-attention" applied to this encoding vector enhances temporal context in transformers by computing associations between pairs of words in the input sequence. We suggest that waves of neural activity traveling across single cortical areas, or multiple regions at the whole-brain scale, could implement a similar encoding principle. By encapsulating recent input history into a single spatial pattern at each moment in time, cortical waves may enable temporal context to be extracted from sequences of sensory inputs, the same computational principle used in transformers.

**Main Text**

*Transformer networks use encoding vectors to capture long-range dependencies*

Cortical mechanisms for spatial context are well established, mediated by the long-range horizontal connections that give rise to non-classical receptive fields[1–4]. The contextual modulations of non-classical receptive fields allow spatial contrasts between inputs to neurons in neighboring cortical columns to be directly encoded into the responses of their classical receptive fields. Equally important, however, is temporal context, which occurs, for instance, when reading words in sentences over a sequence of saccades. Consider these sentences in French and English:

"Le chat à traversé la rue parce qu'il faisait chaud."

and

"The cat crossed the street because it was hot."

Contextual information is indispensable in translating one sentence to the other. In the English sentence, "it" may refer to the cat, the street, or the weather more generally. In the French sentence, by contrast, "il" could not refer to the street but only to either *chat* or the weather. Decisions regarding the referent of "it" are determined by context – whether within the sentence, in the context of a neighboring sentence, or the context of a whole paragraph. Experienced readers parse this effortlessly based on their experience. Language is chock full of these context-dependencies, which can make for surprises, as when Groucho Marx announced, "This morning I shot an elephant in my pajamas. How it got in my pajamas, I don't know."

Over the past twenty years, interest in predicting sequences of words has steadily increased in natural language processing (NLP). With vast amounts of text available online, there was great interest in learning models to predict and generate naturalistic language from this text. Neural networks that could generate sequences were an obvious choice for this task. Networks where model neurons are connected in a dense web, called *recurrent neural networks* (RNNs), have long been known to be effective for generating sequences[5]. RNNs are distinct from feedforward models, such as convolutional neural networks[6,7], where neurons are organized into successive processing layers with no internal, intra-layer connections. Inputs to an RNN affect neurons within the network, which then propagate their activity to other neurons through a dense and loopy web of interconnections (Fig. 1, top left). An RNN receiving words in a sentence as inputs, one by one, can build up an internal state that can, in turn, capture dependencies within a natural language sequence[8]. Various techniques were developed to train RNNs[9–11]. In applying RNNs to sequence prediction for natural language tasks, however, researchers began to realize the difficulties in training RNNs to pick up on long-range dependencies[12–14], which are critical for language prediction, such as with the context-dependent gist a human reader picks up with ease.

To address problems with long-range dependencies, a new mechanism was introduced by Bahdanau, Cho, and Bengio in 2014, allowing an RNN to learn which parts of a source sentence were the most valuable for making correct predictions[15]. This mechanism was called "attention" and had a loose association with the process of human attention to different sensory items. This mechanism allowed the network to "focus" on those pieces of the input sequence that would most effectively drive its internal state to produce the correct prediction. This mechanism for identifying predictively valuable segments proved very effective in helping RNNs learn natural language prediction tasks. Equipped with attention mechanisms, RNNs gained proficiency in sentence-level translation tasks, in part tackling the problem of long-range dependencies. However, a breakthrough in utilizing attention for long-range dependencies was reported in 2017 with the introduction of transformer networks[16]. The main innovation behind transformers was surprising: they used only the attention mechanism and relatively simple feedforward layers to predict the next word. This simplified architecture provided the foundation for advances exhibited in the Generative Pre-Trained transformer (GPT) architecture[17] and led to the current large-language model (LLM) chat agents such as ChatGPT, LLaMa, PaLM2, and Gemini[18].

In a language translation task, the transformer architecture is divided into an Encoder (which, in the above example, would process the sentence in French) and a Decoder (which would output the sentence in English) (Fig. 1, bottom left). The critical step is the self-attention module, where a set of features learned for each word item interacts with features for the other items in the input sequence (Fig. 1, right). The size of this feature vector is called the "embedding dimension." If the feature vector for one word matches another, the two words will have a vital link in the self-attention process. For example, in a given input sequence, "popcorn" and "ribosome" will be less strongly linked than "popcorn" and "movie". Once this process is computed in parallel for all the words in the input sequence, the array of numbers storing the embedding vectors for all words in the input passes into a simple feedforward network. This is the basic function of one Encoder module in the transformer architecture. The self-attention mechanism is repeated many times within a single Encoder. This process is called "multi-headed attention." After several encoding layers with self-attention (Fig. 1), the resulting encoding vector then passes to a multi-layer Decoder, where the relationships the encoding vector has captured with self-attention aid in the correct prediction of the next output. During the training process of the transformer, the connections that make up the self-attention and feedforward modules in the Encoder learn how to create a very high-dimensional encoding vector that can effectively drive the decoder to predict correct output sentences. The long vectors used to encode the inputs, together with the self-attention across the components of the vector, provide a comprehensive context for making predictions. The temporal context of spoken words is represented in a transformer by the spatial context within the encoding vector.

The transformer architecture introduced the idea that the attention mechanism was "all you need" for language prediction tasks. As transformer networks scaled up, their encoding vectors became surprisingly proficient at capturing the long-range dependencies in language that were previously difficult to capture with standard RNNs, which received words sequentially. The breakthrough represented by the transformer is that *the computation itself is simple* - the self-attention mechanism, iterated with feedforward networks, dramatically increased computational efficiency, meaning these networks could be scaled in size and unexpectedly solved many natural language problems. As has become clear in GPT models, these networks can successfully produce coherent pages of text and, in some cases, display impressive generalization and reasoning[19]. The utility of this encoding vector and its focus on capturing the relationships between words in an input sequence is central to these advances in language prediction. Grasping what this encoding vector can teach us about computation more generally could advance our understanding of neural networks, both artificial and biological. In what follows, we focus on experimental data from single regions of sensory cortex to understand how this encoding principle from transformers could be implemented at this scale. In the concluding sections, we discuss how this principle could be implemented by interactions across cortical regions at the whole-brain scale.

*Capturing relationships by encoding a complete input sequence in parallel*

In learning and using context, do brains rely on anything like a transformer? The fundamental insight of the encoding vector is to capture *in parallel* predictively valuable relationships between all the items in an input sequence, rather than handling inputs one by one, as with standard RNNs. Sensory inputs, however, arrive at our brains one spoken word and one eye fixation at a time. This appears to pose a fundamental difficulty for brain systems to use transformer-style contextual information, that is, to capture relationships *in parallel* and to operate on sensory input, which is astronomically high-dimensional and continuously arrives at the periphery. Yet somehow, brains do seem to have solved some version of the broad context problem. Could brain circuits implement an encoding strategy that is similar to that of transformers?

The tactic of encoding many elements in a temporal sequence in parallel may at first appear at odds with our current understanding of sensory processing in the brain. Regions in the sensory cortex contain neurons that respond selectively to the onset of sensory inputs[20]. For example, the orientation of a bar of light may be encoded by the spike rate of an orientation-selective unit in primary visual cortex (V1), the tone of a sound may be encoded by the rate of a frequency-selective neuron in primary auditory cortex (A1), or (to take an example from a cognitive system) the position of a rodent during navigation may be encoded by the spike rate of "place cells" in the hippocampus. Hubel and Wiesel, in their pioneering work on neuronal selectivity in the visual cortex[21,22], established a model in which the input entirely drives the sensory encoding and where just-now events in its receptive field determine an individual neuron's response properties. In this model of sensory encoding, trial-to-trial fluctuations that deviate from the average response expected from the receptive field are thought to be a product of noise[23] or to represent uncertainty[24,25]. Lateral interactions due to horizontal connections are known to influence selectivity, specifically through the non-classical receptive field[1–4]. These effects have been proposed to mediate specific computations in the visual system, such as contour completion[26]. Feedback projections[27] from higher cortical areas have also been proposed to provide context for incoming sensory inputs or to impose bias on incoming sensory input[28]. This feedback is composed of efference copy from motor commands in addition to sensory information[29,30]. These circuit features add additional computations for spatial context into the feedforward processing model of Hubel and Wiesel; at the same time, however, they do not explain how cortical circuits could take advantage of activity generated by inputs from the recent past, in turn enabling these circuits to perform computations with temporal context.

On the other hand, the powerful transformer strategy of encoding entire input sequences in parallel, along with their predictive relationships by virtue of a sizable encoding vector, appears ill-suited to biological neural networks as characterized by the classical framework of sensory function. At first blush, the transformer strategy seems beyond the brain's reach, assuming that neuronal encoding remains a fixed function of single input features, such as visual orientation or auditory pitch at one moment of time. That assumption, we suggest, may benefit from another look in the light of new recordings from arrays of electrodes.

Recent research has demonstrated that rather than being based solely on just-now features of sensory input that are currently present, the selectivity of single neurons may take future, predicted features as well as past features into account. One recent study of place cells in bats noticed that, by shifting the present position of the animal forward in time in the data analysis, hippocampal place fields became sharper, and new, well-formed place fields became apparent[31]. This result suggests that, especially at the high speeds flown by bats relative to place field size in the hippocampus, selectivity may be enhanced by future, anticipated inputs rather than restricted to present input stimuli. Anticipatory responses to moving stimuli have also been observed in the visual system, in the peripheral circuits of the salamander and rabbit retina[32,] and, more recently, in monkey V1[33]. These results, which were obtained by averaging across trials, indicate that in contrast to simply reflecting present sensory input, the sensory and cognitive systems' maps may play more dynamic roles in neural computation. If that is the case, the question is this: could the highly structured encoding that occurs in transformers to handle contextual features across time be enabled by the dynamics of these circuits on a trial-by-trial level?

*Waves in single regions of visual cortex: parallel encoding of the recent past*

A critical computational insight of the transformer architecture is to encode the words in an input sequence in parallel, in a highly structured encoding that allows extracting meaningful relationships. Regarding the visual system, first consider a series of points of light presented briefly. The key circuit element would be a way to link the activity patterns evoked by each stimulus, even after the initial activity pattern has subsided, to generate predictively valuable signals. How might neural circuitry be organized to achieve transformer-like richness?

Studies using large-scale optical imaging techniques and multielectrode array recordings in non-human primates have demonstrated that small visual stimuli drive waves of activity that propagate from the input point across the visual cortex[34,35]. These waves propagate at the same speed as the unmyelinated long-range horizontal fibers that connect neurons across cortical areas[36], traveling over a substantial portion of the map of visual space in tens of milliseconds. These long-range fibers, which project many millimeters to connect neurons across an individual cortical region[26], are thus a candidate network mechanism underlying waves in single cortical regions, such as V1.

How are such waves generated? Waves occur quite generally in networks of rate-coded neurons with local interactions[37–39]. Waves in these networks are "dense", in the sense that each neuron participates in the wave as it passes across the network. Spiking neural networks, however, can admit a distinct activity state: "sparse" waves, where only a few neurons spike as the wave travels across the network, primarily when known distance-dependent axonal conduction delays are included[40]. Networks with sparse waves can match the activity observed in experimental recordings of visual cortex. Waves in this model propagate with the distribution of speeds observed in experiments, with activity at a local scale remaining consistent with the low-rate, decorrelated "asynchronous-irregular" activity regime. Evidence from both models and experimental recordings indicates that spontaneous and stimulus-evoked waves involve the contribution of many synapses in coordination, rather than strong monosynaptic connections at spike initiation zones[41]. Further, in experimental recordings, waves modulate neural excitability and thus the responses to incoming inputs[42,43]. These large-scale spiking network models also reveal that the local balance of excitation and inhibition of neurons is modulated as the waves pass through local circuits, which in turn modulates excitability to incoming stimuli[40].

Experimental observations indicate that waves evoked in the awake state do not cross the boundaries between different cortical areas[34], in contrast to those in anesthetized animals that do[44,45]. This restriction suggests that waves occurring during normal, waking visual processing respect the retinotopic maps in individual regions of the visual system. With neurons thus organized, the waves could yield structured spatiotemporal patterns in response to sequences of brief input stimuli. In both recordings and models, another important feature of these waves is that they are sparse: when a wave passes over a local patch of cortex, only a tiny fraction (< 1%) of the neurons spike. This profile contrasts with the dense waves that occur, for example, during epileptic seizures. Unlike dense waves, sparse waves propagate across single cortical regions along long-range horizontal fibers, modulating but not completely overwhelming the feedforward input.

These experimental and modeling results raise the possibility that stimulus-evoked waves are not pointless doodads, but may play a significant computational role in sensory processing. In this case, however, what computation could this be? Waves of neural activity traveling over the retinotopic map seem at first inconsistent with the standard framework for sensory processing. Within this canonical framework, visual system models generally consider feedforward inputs from the retina, with precise retinotopic projections from one layer of neurons to the next, to process incoming visual inputs through successively elaborated receptive field selectivity[46].

The objection is this: if a single-point stimulus can evoke a wave that travels over a large part of an individual visual area, such a wave could disrupt the processing of other sensory stimuli as it propagates. Consequently, stimulus-evoked waves appear incompatible with the classical conception of precise retinotopic maps and retinotopic projections. Nevertheless, a closer look suggests an alternative in which mixing information across space and time has computational advantages.

Waves traveling across single regions of cortex can provide *temporal context*[47] and can also convert temporal information into spatial codes[48]. We have previously introduced a potential computational role for waves traveling over topographic maps in single regions of sensory cortex[35]: waves traveling radially outward from the point of input can encode both *where* (in retinotopic space) and *when* a stimulus occurred. To take a simple example, with a small, punctate input that evokes a wave (Fig. 2, top), a decoder could tell *where* the input occurred by using the center point of the wave on the retinotopic map and *when* it occurred by using the distance from the center and the fact that these waves travel at a specific range of speeds. *Where* and *when* are thus encoded in the wave pattern. In a case with multiple stimuli, such as multiple spots presented in a sequence, both the sequence of stimulus positions and their onset times could be decoded from the spatiotemporal interference pattern in the membrane potentials of the neural population (Fig. 2, bottom). In this way, waves could provide a mechanism for the sensory cortex to encode stimuli in the recent past in a highly structured manner that enables extracting meaningful relationships across space and time.

As waves of activity spread laterally within the cortex, they influence the spiking activity of neighboring neurons after a delay caused by conduction through unmyelinated long-range horizontal axons (Fig. 3a). As the wave progresses through the tissue, it influences the spiking activity of more distant neurons after further time delays, as visualized in Fig. 3b as an expanding spacetime cone. This diagram abstracts the causal structure of all inputs that can influence a spike in a single neuron, which can then influence other neurons downstream. In a natural scene, many neurons will be activated, potentially creating complex interference patterns between all the sparse, expanding waves. This is reminiscent of a hologram formed by spatial interference fringes, which contain all the information needed to recreate a 3D object when illuminated by a coherent light source. In the cortex, spatial input is mixed with temporal delays to create a dynamic *spacetime representation* containing information needed to recover the spatial and temporal history of the sensory inputs.

*Spacetime representations may be useful for processing inputs across topographic maps*

The neural activity underlying population codes is traditionally viewed as a *separable* function of space and time, i.e. *P(x,t) = F(x)G(t)*, where *P(x,t)* is a function describing the profile of neural population activity and the other two terms are functions of only space and only time, respectively. Here, "space" refers to cortex or, equivalently, to sensory space when the cortical area is organized into a topographic map. In contrast, waves indicate that the neural activity underlying population codes may be space-time *nonseparable* at the moment-by-moment level, such that the function *P(x,t)* cannot be decomposed into two independent functions for space and time. In this case, neural population activity does not represent information at a single moment in time, but instead can also contain activity from the recent past, in the form of waves propagating over the topographic map: *P(x-vt)* where *v* is the velocity of the wave.

How could this "mixing" of information possibly be useful in cortical computation? As noted above, a key feature for computation may be that waves provide a mechanism to encode stimuli in the recent past in a structured manner, as continuous spatiotemporal structures traveling over the topographic sensory maps. A theoretical model has shown that waves can indeed enact a conjunctive encoding of *where* and *when* a stimulus occurred and, in addition, can drive short-term predictions of incoming sensory inputs[49]. The recurrent network model driving these short-term predictions incorporates the main architectural features of single regions in cortex – local connectivity and distant-dependent time delays. Short-term predictions are possible in this recurrent network when the strength of feedforward and recurrent inputs is approximately matched, in general agreement with the ratio of feedforward and recurrent input to individual neurons in V1 (measured under anesthesia)[50,51]. When connections in the recurrent network are randomly shuffled, the network does not produce accurate predictions, even after retraining. These results demonstrate that, when RNNs follow the basic architectural features found in visual cortex, waves may provide a unique way to embed short-term predictions onto the retinotopic map, in a more highly structured form than general patterns of activity produced by networks of randomly connected neurons.

Recent evidence from training RNNs has also begun to suggest further roles waves could play in neural computation and prediction more generally. Training RNNs to predict sequences naturally results in recurrent weight matrices with Mexican hat connectivity, which supports waves[52] (see section "*Potential implementations of self-attention in cortical networks*" below). Comparing locally connected RNNs that generate waves with randomly connected RNNs that do not generate waves showed that wave-generating networks could be trained to perform more complex sequence learning tasks more easily, with training almost two orders of magnitude faster than randomly connected networks[53]. Finally, waves in RNNs can drive elementary computer vision tasks such as image segmentation[54]. These results were inspired by the main organizational features in single cortical areas (local recurrent connections and distance-dependent time delays) and the principles learned from transformers.

*Waves and transformers: bringing the encodings together*

The potential similarity between transformers and waves is that they may be tapping into the same computational principle, albeit with somewhat different physical mechanisms: by processing inputs in parallel, using a highly structured encoding, transformer networks and cortical waves may enable extracting meaningful relationships from these sequences. In the case of the transformer, the long encoding vector contains the attention mechanism that enables capturing the long-range dependencies critical for natural language processing. In the case of waves in the visual cortex, the highly structured spatiotemporal patterns, earlier tagged as sparse, may enable encoding temporal relationships directly onto populations of neurons over the retinotopic map, facilitating flexible storage of the recent past in a way that enables extracting the temporal relationships from the spatial map.

This potential similarity between the computational principle underlying both waves and transformers may explain the function of waves in single regions of the visual cortex. Since the introduction of the feedforward model of the visual system by Hubel and Wiesel[21,22], and its refinement through successive network implementations[46,55], it has often been implicitly assumed that the visual cortex contains a veridical image of sensory input, albeit filtered in some way by the receptive field selectivity in each area. *Waves in single regions of visual cortex, however, indicate that input encoding in the visual system may be much more sophisticated, since local populations of neurons can influence networks far across the retinotopic map in a highly structured manner.* Encoding long input sequences in parallel provides transformers with an advanced capacity to extract meaningful relationships in natural sensory input. This stunning but conceptually simple achievement suggests that nervous systems could conveniently implement roughly comparable encoding to extract relationships across input sequences. Although the complex activity patterns of the visual system – from spontaneous activity in the absence of visual input, to variable neural responses to identical simple stimuli, and finally to the dynamics in response to naturalistic inputs that are difficult to explain from selectivity estimated from simple isolated stimuli[56–58] – may have at first been taken to be meaningless fluctuations, it is possible that these fluctuations are not mere noise. Instead, they may reflect computations that extract meaningful relationships from the continuous stream of visual input and create short-term predictions of incoming stimuli. Although we have largely focused on neural circuit phenomena here, experiments linking circuits to behavior, for example with expectation effects such as priming, show clear speed and accuracy benefits in making accurate predictions of upcoming sensory inputs, such as auditory anticipations driven by experience[59,60].

*Potential implementations of self-attention in cortical networks*

As shown in Fig. 1, the input to each layer projects to self-attention, which then is combined with the feedforward projection. So-called self-attention is a relatively recent addition to deep feedforward networks. The foundational paper for LLMs was entitled "Attention is All You Need," emphasizing its importance[16]. Without self-attention, a transformer would be a conventional feedforward network with limited capabilities. Here is how GPT-4 described self-attention: "Imagine you're reading a book and come across a sentence that refers to something mentioned a few pages back. You might flip back to remind yourself. Self-attention allows the model to look at other words in the sentence to better understand the current word." This is different from how "attention" is used in neuroscience, which typically is focused on single sensory items. Nonetheless, "self-attention" could be considered a generalization of attention that links items across time.

How could "self-attention" be implemented in cortex, whose highly recurrent architecture differs from the matrix self-attention and feedforward layers in transformers? State space models have recently emerged as a more efficient alternative to transformers, specifically by replacing the matrix self-attention mechanism with a much more computationally efficient linear convolution[61–64]. A state-space model (SSM) takes in temporal sequences of vector inputs and transforms them into sequences of vector outputs. SSMs are already well established in the motor system, and in particular the motor cortex, where muscles are controlled to follow dynamical trajectories[65]. One of the simplest SSMs that is amenable to analysis is given by a first-order linear differential equation (Fig. 4). In this system, the vector **x** can represent the state of *N* neurons in a recurrent network, whose connections are defined by **A**. The system is responding to an input specified in vector **u** by generating an output **y**. From this perspective, this simple SSM is similar to a standard RNN with linear instead of nonlinear activation function[66].

For the SSMs that are used to implement mechanisms similar to self-attention in transformers, **A** has a special form, called a Toeplitz matrix, where the value of the parameters on each diagonal from left to right is a constant[67]. Mexican hat connectivity is an example of a Toeplitz matrix. We have recently developed a mathematical theory that links the structure of these matrices (more specifically, circulant matrices, which are a special kind of Toeplitz matrix[68]) to waves in network systems, even when the dynamics are nonlinear (n.b. this mathematical approach also generalizes to other types of connectivity patterns, including random graphs)[69,70]. If the dynamics of individual nodes are nonlinear, then relating the pattern of connections in a connectivity matrix to the network dynamics is a difficult problem. Fortunately, the new mathematical tools smoothly handle this complexity, enabling the emerging theory to explain the connection between traveling waves and the structure of SSMs that are now used for self-attention in machine learning tasks. It remains to be seen whether future theoretical work can identify significant links between the spatiotemporal dynamics of neural populations and the dynamical activity patterns that result from these SSMs (and recurrent networks more generally).

*Future Developments*

Many open questions regarding traveling waves and their functions invite a range of conjectures. For example, are "self-attentional" structures learned during childhood? For another: how could events originating from separate cortical areas be linked together? Scaling up spacetime cortical codes more broadly requires distant cortical regions to interact on a larger scale. Waves at the whole-brain scale have been observed with electroencephalogram (EEG) and magnetoencephalogram (MEG) recordings in humans during wake[71–76] and sleep[77]. While signal blurring poses a significant challenge to quantifying spatiotemporal dynamics with these non-invasive techniques, waves have also been studied with intracranial recordings in humans during wake[78–80] and sleep[81], and in rodents, where an entire hemisphere of cortex can be imaged simultaneously with optical techniques[82–86]. It is also tempting to wonder whether major brain regions that have reciprocal loops with the cortex – the basal ganglia and the cerebellum, for example – might serve to coordinate interactions between distant cortical regions, among other possible roles. The reciprocal loops between the cortex and the basal ganglia are topographically organized[87,88] (Fig. 5). The basal ganglia[89] are known to be involved in learning and generating sequences of actions to achieve goals and could be a site for self-attention. Regarding the cerebellum, temporal context is essential for fast coarticulation in speech and transformer-style self-attention could facilitate coordinating muscular contractions by extending motor representations over time as spatial activity patterns.

And surely this question arises: Why are there large differences between traveling waves characteristic of sleep, which likely involve thalamocortical loops and intracortical connections[90], and those typical of the awake state? Analyses of spatiotemporal dynamics across different cortical states, at different states of wake and anesthesia[91], could provide critical insights into how waves are shaped by changing interactions between thalamus and cortex, in addition to other subcortical structures[92–94]. Finally, how does spontaneous cortical activity – where waves occur in individual cortical regions[45,95] and at the whole brain scale[82,83] – interact with stimulus-driven waves, and what are the implications for the computations discussed here? For example, by continually refreshing recent inputs, spontaneous waves could be extending working memory.

Moving forward, close interaction between theory and experiment will be critical for testing these ideas with specific, model-driven predictions. Technological improvements are rapidly advancing the scale of neural recordings and new analytic techniques can be applied to test the predictions of spacetime coding. One key prediction emerging from this framework is that, as strong feedforward sensory input arrives at a cortical neuron, its response will be mixed with information about inputs from distant spatial locations at previous times (Fig. 3b). This can be experimentally tested by reconstructing previous sensory inputs from current activity in large populations of neurons. Second, because the spacetime population code is spatially distributed, the information extracted across the cortex should be highly overlapping. Some signals, such as efference copy motor signals and neuromodulatory signals, are broadly represented. Finally, membrane potentials recorded using voltage-sensitive dyes should have more information than spikes, which are sparse, and there should be more information in the relative timing of spikes in neural populations about past sensory inputs compared with firing rates.

*Discussion*

Waves traveling across single cortical regions are ubiquitous in the cerebral cortex, as observed in sensory[34,43,96], motor[42,97–99], and prefrontal[100,101] regions during normal sensory processing and behavior, as well as in the hippocampus[102–104] and basal ganglia[105]. Electromagnetic and sound waves carry delayed information across space, and neural waves carry delayed information across cortical space.

We have focused here on waves in single regions of cortex. A possible function for stimulus-driven waves traveling over topographic maps is to provide temporal context for sequences of sensory inputs. In the prefrontal cortex this spacetime code could integrate words over a longer timescale, for example during an hour-long lecture, a form of long-term working memory[106]. The waves we discuss mix old information with new information delivered by feed-forward inputs to create a new type of spacetime population code. This form of encoding has computational advantages similar to those found in the transformer architecture of LLMs, which map temporal sequences into a long input vector. Evolution may have found an alternative method to achieve the same functionality, taking advantage of cortical dynamics in recurrent networks.

Throughout the biological world, evolution has repeatedly exploited the physics of oscillators to extensively use waves in systems on a wide range of time scales, from the rotation of flagella to whisking, digesting, egg-laying, and swimming[107]. We hypothesize that another evolutionary adaptation deploys waves of neural activity specially suited to sparse spiking dynamics in the cortex in mammalian and in lower vertebrate brains to support spacetime coding[47].

Population coding by waves traveling over topographic maps may not be as intuitive initially as the traditional conceptual framework for coding with separable receptive fields in sensory maps. Both, however, might be relevant to understanding neuronal function, specifically as multi-tier levels of description of sensory systems. Receptive fields are measured under carefully controlled conditions, repeating nearly identical sensory stimuli, and then averaging over neural responses to many presentations. Receptive fields thus capture information about a neuron's responses *on average* to features of sensory stimuli, with trial-to-trial fluctuations about this average thought to be a product of noise. This framework for neural coding has been highly successful in understanding responses to repeated visual stimuli and in understanding the elaboration of neuronal selectivity across the visual system. However, when faced with an incoming stream of complex whole-field visual inputs, extracting meaningful relationships across time, as in transformer networks, using spacetime coding may give sensory systems an important advantage in predicting upcoming inputs and preparing behavioral responses.

**Acknowledgements**

The authors thank Andrew Keller and Max Welling for inspiring discussions on waves in trained recurrent neural networks and Arjun Karuvally and Hava Siegelmann for discussing their models showing that traveling waves in recurrent networks can implement variable binding, an essential computational primitive. April Benasich helped us analyze sleep spindle waves in infants and how these waves could influence brain development. Luisa Liboni provided helpful comments on the manuscript. This work was supported by grants from ONR N000014-23-1-2069, NIH DA055668-01, MH132644-01, NSF NeuroNex DBI-2014862, the Swartz Foundation, BrainsCAN at Western University through the Canada First Research Excellence Fund (CFREF), NSF/CIHR NeuroNex DBI-2015276, the Natural Sciences and Engineering Research Council of Canada (NSERC), the Western Academy for Advanced Research, NIH U01-NS131914, and NIH R01-EY028723.

**Declaration of interests:** The authors declare no competing interests.

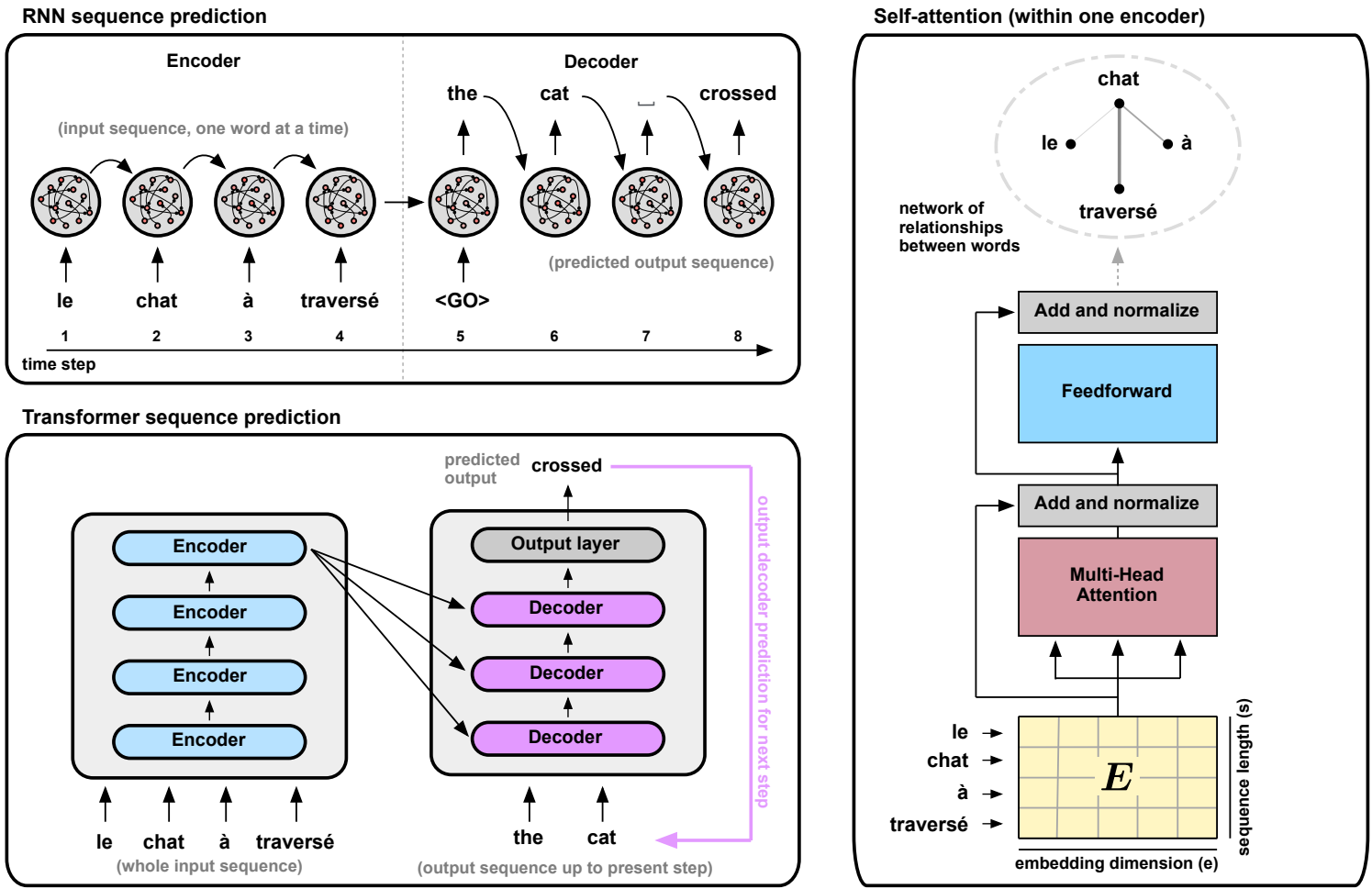

**Figure 1. Recurrent neural network architecture, transformer networks, and self-attention in language prediction tasks. (Top left)** RNNs for language prediction tasks take each word in a sequence as input, one at a time. The inputs are processed by the RNN, whose state passes from time step to time step (horizontal arrows) in order to build up a representation of the sequence. After the entire input sequence is fed into the RNN, a "go" signal cues the network to generate the output sequence, again one word at a time. Each generated output word is fed back into the RNN to recursively generate the output sequence. **(Bottom left)** Instead of taking each input one at a time, transformers take in the whole input sequence, which is processed through a series of Encoders. GPT-4 has a context length of 128k tokens (about 240 pages at 400 words per page). The output of the last Encoder is then an input to the Attention mechanism in the Decoder modules. The output of the complete Encoder-Decoder is the predicted next word in the sequence. This prediction is then appended to the input to the decoder to start the prediction for the next step. **(Right)** Within a single layer of the Encoder and Decoder, the sequence encoding (E) is passed to a multi-head attention module. The result of this calculation is the self-attention score, which is added to its input and passed on to a traditional feedforward layer. This self-attention mechanism enables the data-driven discovery of the network of relationships between words in the input sequence (top). Note that the input is added to Multi-Head Attention in the Add and Normalize box, which externalizes it as a parallel module.

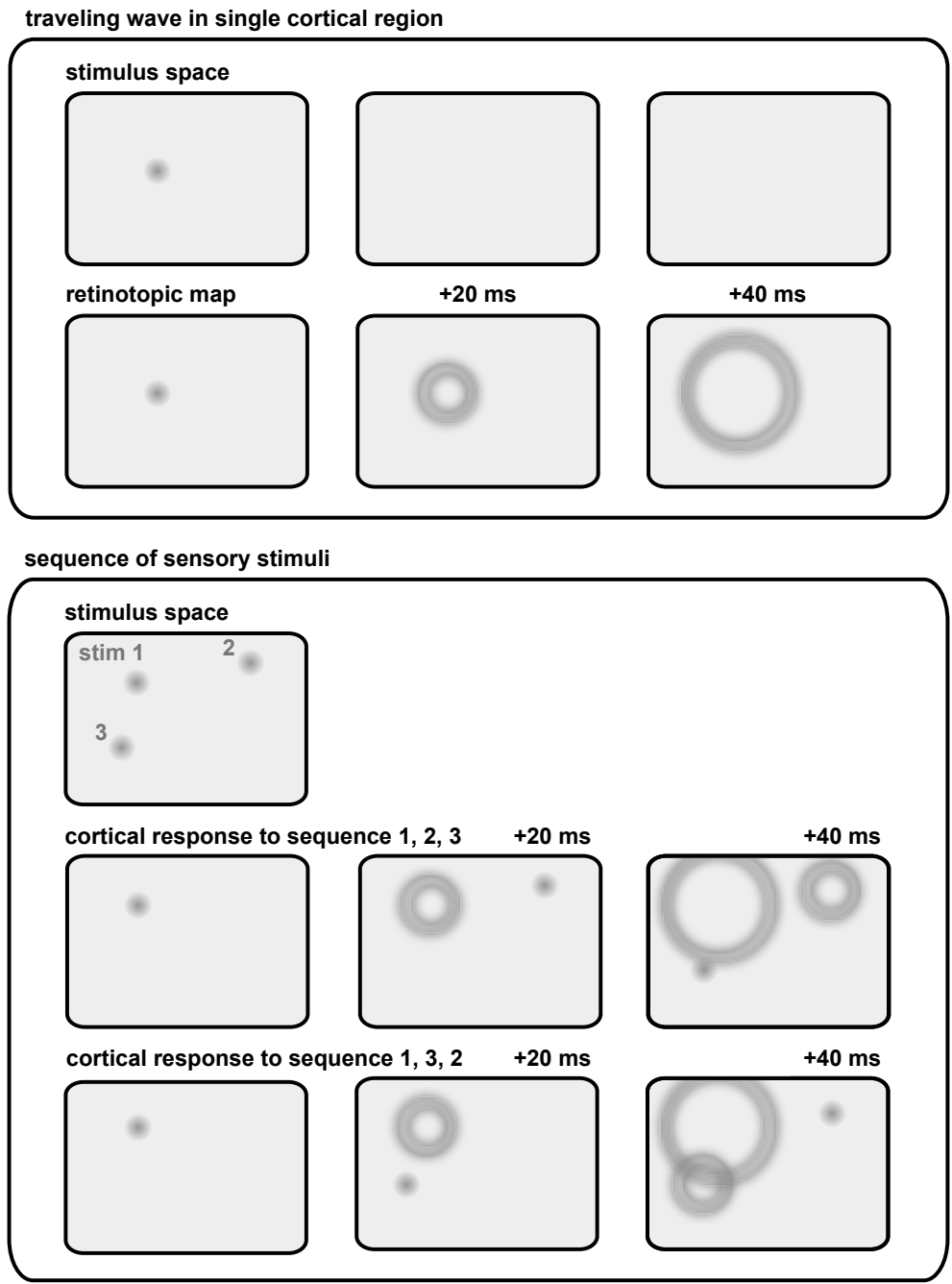

**Figure 2. Schematic of stimulus-evoked waves in single regions of cortex and responses to multiple stimuli. (top box)** Analysis of optical imaging data in awake, behaving primates has revealed that small, punctate visual stimuli (stimulus space, top row) can create waves of activity that propagate outward from the point of feedforward input[34] (retinotopic map), similar to ripples in a pond created by dropping a pebble. **(bottom box)** In the case of three visual stimuli (stimulus space, top row), a specific temporal order of presentation (stimulus 1, then 2, then 3) can create one pattern of waves (cortical response to sequence 1, 2, 3), while another order of presentation (stimulus 1, then 3, then 2) can create a different spatiotemporal pattern (cortical response to sequence 1, 3, 2).

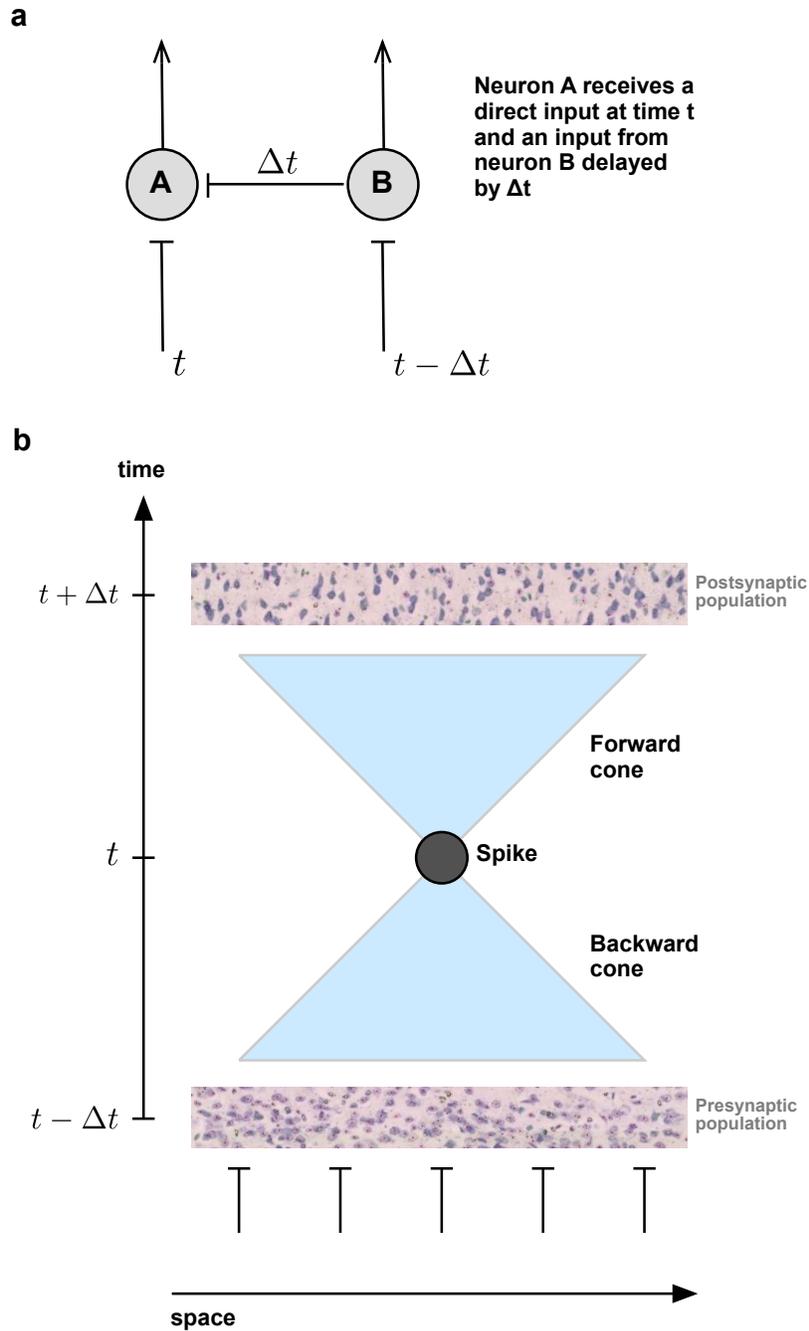

**Figure 3. Time delays between laterally interacting neurons create a spacetime population code.**
**(a)** Neuron A receives a direct input at time *t* and an input from neuron B delayed by *Δt*. **(b)** The response of a spiking neuron (dark gray circle at the intersection of the two blue triangles) is influenced by the activity of all the interacting neurons in the backward spacetime cone (blue triangle from *t - Δt* to *t*), as structured by the temporal delays in the network. The spike of the neuron at time *t* influences, in turn, a population of interacting neurons within the forward spacetime cone (blue triangle from *t* to *t + Δt*). The backward cone extends back in time to include all inputs to the central neuron and the forward cone extends forward to all neurons in its projective field.

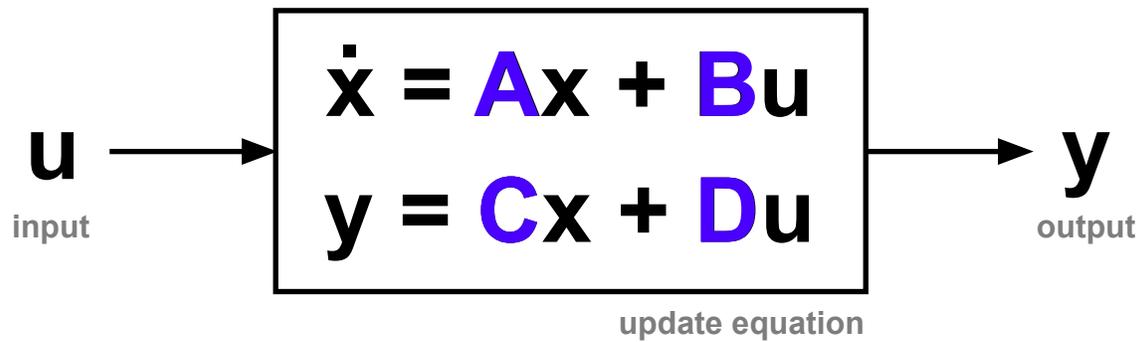

**Figure 4. Input-output relationship and equations for state-space models (SSMs).** In this figure, **u** is the input vector and **y** is the output vector. The vector **x** is an intermediate dynamical variable and **x** dot is the first derivative. The matrices **A**, **B**, **C** and **D** control the dynamics. In control theory, **A** is a model of the dynamical system and in the cortex **A** is the connectivity matrix, which generally resembles a mexican hat with excitatory connections between nearby pyramidal neurons and inhibitory influence on more distant neurons.

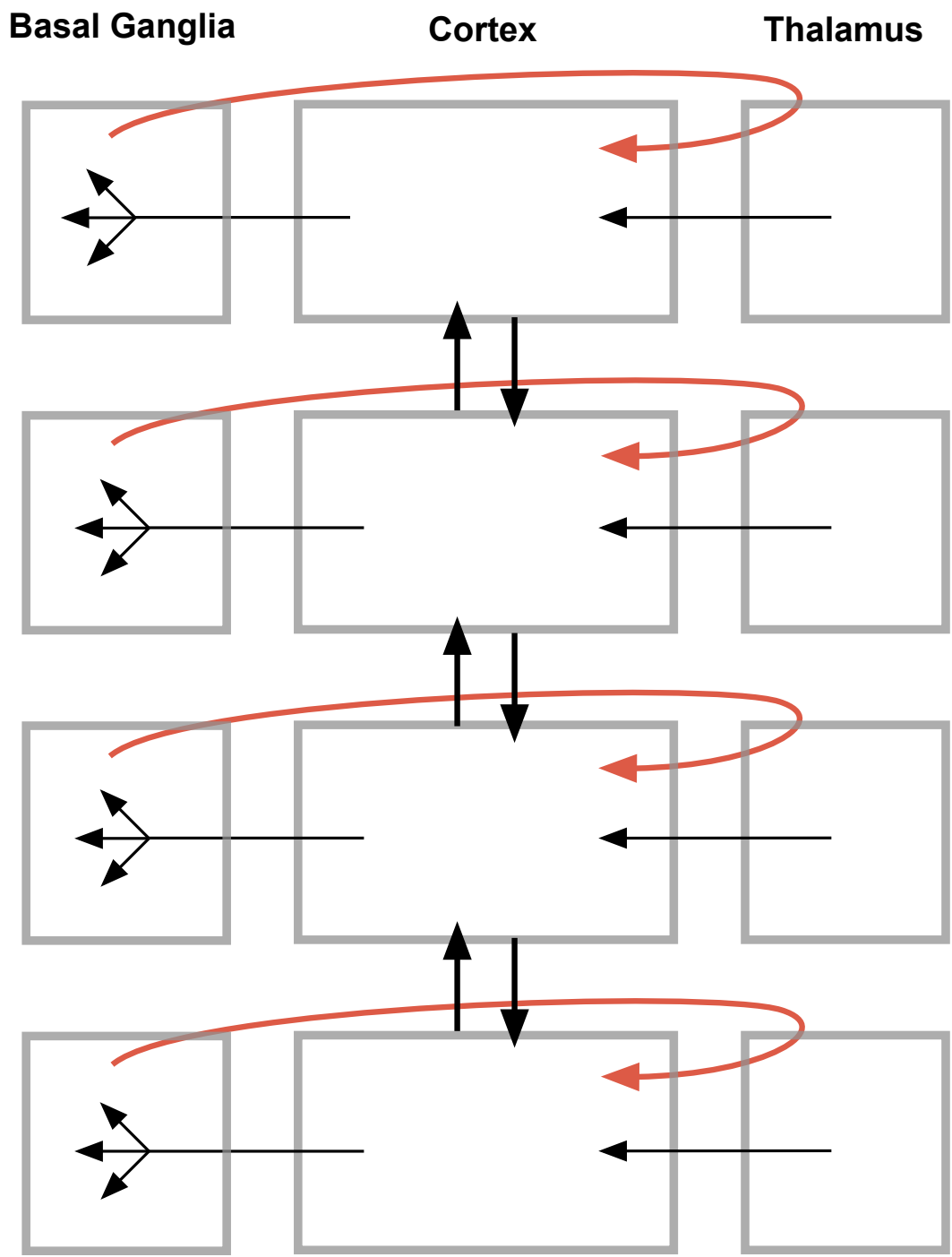

**Figure 5. Schematic diagram of the loops between the basal ganglia and the cortex.** Cortical areas project topographically to the basal ganglia, which then feedback topographically to the cortex through the thalamus. Compare this with the self-attention box in Fig. 1. Cortical hierarchies are found in sensory cortex, motor cortex and the prefrontal cortex. Associations between input to the basal ganglia can be learned through dopamine neurons, which carry reward prediction signals. The cortex receives inputs from the thalamus, similar to the encoder inputs that the decoder receives in a transformer.